\documentclass[journal]{IEEEtai}

\usepackage[colorlinks,urlcolor=blue,linkcolor=blue,citecolor=blue]{hyperref}

\usepackage{color,array}

\usepackage{graphicx}

\usepackage{booktabs}
\usepackage{multirow}
\usepackage{url}
\usepackage{subcaption}
\usepackage{float}
\usepackage{amssymb}
\usepackage{amsmath}
\usepackage{stfloats}

\begin{document}

\title{SCAN: Visual Explanations with Self-Confidence and Analysis Networks}



\author{Gwanghee Lee$^\dagger$, Sungyoon Jeong$^\dagger$, and Kyoungson Jhang, \IEEEmembership{Member, IEEE}
\thanks{$\dagger$ They contributed equally.}
\thanks{(Corresponding author: Kyoungson Jhang.)}
\thanks{G. Lee and K. Jhang are with the Department of Computer Science and Engineering, Chungnam National University, Daejeon 34134, Republic of Korea (e-mail: manggu251@o.cnu.ac.kr, sun@cnu.ac.kr).}
\thanks{S. Jeong is with the Department of Artificial Intelligence, Chungnam National University, Daejeon 34134, Republic of Korea (e-mail: jungbug04@g.cnu.ac.kr).}
}

\markboth{Journal of IEEE Transactions on Artificial Intelligence, Vol. 00, No. 0, Month 2020}
{Lee \MakeLowercase{\textit{et al.}}: SCAN: Visual Explanations with Self-Confidence and Analysis Networks}

\maketitle

\begin{abstract}
Explainable AI (XAI) has become essential in computer vision to make the decision-making processes of deep learning models transparent. However, current visual explanation (XAI) methods face a critical trade-off between the high fidelity of architecture-specific methods and the broad applicability of universal ones. This often results in abstract or fragmented explanations and makes it difficult to compare explanatory power across diverse model families, such as CNNs and Transformers.
This paper introduces the Self-Confidence and Analysis Networks (SCAN), a novel universal framework that overcomes these limitations for both convolutional neural network and transformer architectures. SCAN utilizes an AutoEncoder-based approach to reconstruct features from a model's intermediate layers. Guided by the Information Bottleneck principle, it generates a high-resolution Self-Confidence Map that identifies information-rich regions.
Extensive experiments on diverse architectures and datasets demonstrate that SCAN consistently achieves outstanding performance on various quantitative metrics such as AUC-D, Negative AUC, Drop\%, and Win\%. Qualitatively, it produces significantly clearer, object-focused explanations than existing methods. By providing a unified framework that is both architecturally universal and highly faithful, SCAN enhances model transparency and offers a more reliable tool for understanding the decision-making processes of complex neural networks.
\end{abstract}


\begin{IEEEImpStatement}
Deep learning models, particularly Convolutional Neural Networks(CNNs) and Transformers, are increasingly deployed in high-stakes computer vision applications such as autonomous driving and medical diagnosis. However, a critical barrier to their wider adoption is the lack of a unified interpretability framework; existing Explainable AI(XAI) methods force a trade-off between the high fidelity of architecture-specific tools and the broad applicability of model-agnostic ones. This paper addresses this fragmentation by introducing the Self-Confidence and Analysis Networks(SCAN), a universal framework capable of providing high-fidelity visual explanations across diverse architectures.
The proposed method utilizes an Information Bottleneck-guided reconstruction mechanism to visualize decision-making processes transparently. Extensive experiments on ImageNet demonstrate that SCAN bridges the gap between universality and fidelity, achieving an AUC-D score of 36.87\%, which is competitive with state-of-the-art architecture-specific methods. Notably, in terms of faithfulness, SCAN reduced the Drop\% by 20.54 percentage points compared to the 'Explainability' method, providing significantly more precise and object-focused explanations. By offering a standardized tool for comparing explanatory power across different model families, this work facilitates more rigorous evaluation of AI reliability and supports the development of trustworthy AI systems in safety-critical domains.
\end{IEEEImpStatement}





\section{Introduction}
\IEEEPARstart{I}{n} recent years, increasing attention has been directed toward making the decision-making processes of deep learning models more transparent, particularly in computer vision, where explainable AI(XAI) has become essential for clarifying how models analyze images and generate predictions~\cite{zou2022ensemble}. Such explanations are vital for evaluating model robustness~\cite{7, kuppa2021adversarial}, countering adversarial attacks~\cite{8,9, liao2024apr}, improving datasets, and optimizing neural networks~\cite{1,2,3,4,5,6}.

The mechanisms of existing methods addressing this challenge are broadly divided into two streams. The first includes universal(or perturbation-based) methods that are model-agnostic, such as LIME~\cite{lime} and RISE~\cite{rise}. The other stream consists of methods highly specialized for specific architectures, such as GradCAM~\cite{exp,11} and LayerCAM~\cite{13} for CNNs, and Explainability~\cite{exp} or Rollout~\cite{roll} for Transformers~\cite{vit, dino, 18, deit}.

However, these approaches have clear limitations. The explanatory power of universal, perturbation-based methods is often significantly lower. Conversely, while architecture-specific methods excel in explanatory ability, they are strongly dependent on their respective architectures, making it impossible to compare explanatory power across different model families, like CNNs and Transformers. Furthermore, the explanations generated by these methods often suffer from ambiguous feature boundaries or abstract regional partitioning, which can lead to misinterpretations of the neural network's actual operations.

To resolve these issues, we utilize a new mechanism based on reconstruction. Our approach is founded on the principle that feature maps in the intermediate layers of a model already retain semantic features~\cite{27,28,29}. We train a model to reconstruct these encoded representations back into the original image space.

While GradCAM is a representative existing work that similarly utilizes feature maps, it only uses the feature map from the last layer. In contrast, the intermediate layer feature maps we aim to leverage are inherently difficult to interpret semantically. We address this challenge by filtering these uninterpretable feature maps using a gradient map as a mask, and then restore and analyze them based on Information Bottleneck(IB) theory.

Overall, this paper proposes a new visual explanation framework named the Self-Confidence and Analysis Networks(SCAN). SCAN is a universal methodology that extracts feature maps from intermediate layers regardless of architecture(both CNN and Transformer), combines them with a gradient mask, and reconstructs them to visualize the core information related to the target class.

The proposed SCAN framework leverages the changes in reconstruction error patterns that arise during this process. This allows the network to visualize the regions where error differences are clearly anticipated—that is, the key features driving the model's decisions—aiming to provide a more precise analysis than existing methods.

\section{Related Works}
Existing research in visual explanations for deep learning can be fundamentally categorized into two primary streams based on their technical approach : (1) model-agnostic perturbation-based methodologies and (2) model-internal methodologies that leverage specific architectural gradients or activations. This section provides a comprehensive review of these approaches, delineating their respective technical constraints to contextualize the necessity of the proposed SCAN framework.

\subsection{Model-Agnostic Perturbation-Based Methods}

The first stream in XAI, perturbation-based methods, are fundamentally model-agnostic, as they do not depend on the model's internal structure. This approach identifies important input regions by perturbing or removing specific parts of the input and observing the resulting changes in the output. Prominent examples include LIME~\cite{lime}, which trains a simple, interpretable surrogate model on locally perturbed data, and RISE~\cite{rise}, which generates a saliency map by analyzing model predictions against a set of randomly generated occlusion masks.

\subsection{Architecture-Specific Methods}

The second stream, model-internal methods, leverages the model's internal mechanics, such as gradients or attention scores. This approach often yields more powerful explanations but is typically specialized for specific architectures.

For CNNs, gradient-based methods are dominant~\cite{11,xcam,12,13}. GradCAM~\cite{11} became a standard by using gradients from the final convolutional layer to weight feature maps. This was later refined by methods like GradCAM++~\cite{12} to improve explanations for multiple object instances, and LayerCAM~\cite{13} to produce richer explanations by integrating feature maps from multiple layers.

For Transformers, attention-based methods naturally emerged. Early techniques like Raw Attention~\cite{exp} and Rollout~\cite{roll} aggregate attention scores across layers to visualize information flow. More advanced techniques, such as Explainability~\cite{exp}, were developed to propagate relevance scores using principles like LRP~\cite{lrp, plrp}. Furthermore, Attention Guided CAM~\cite{leem2024attention} enhances this by adopting sigmoid normalization to mitigate peak intensity noise, enabling more precise localization of full and multiple object instances.

\subsection{The Research Gap: Fidelity vs. Universality}

Despite these advances, existing methods face a critical trade-off that defines the specific problem we address: the fundamental conflict between the high fidelity of architecture-specific methods and the broad applicability of universal ones.

This conflict manifests as clear limitations for each approach. The universal, perturbation-based methods(LIME, RISE) offer flexibility but often suffer from significantly less explanatory power. In contrast, architecture-specific methods suffer from critical constraints. CAM-based explanations for CNNs(GradCAM, LayerCAM) are fundamentally limited by their strong dependency on the underlying convolutional architecture, which hinders their universal applicability. Concurrently, attention-based explanations for Transformers(Raw Attention, Rollout) tend to be class-agnostic. They often reflect the input data's inherent saliency rather than providing specific evidence for a given class prediction.

Therefore, a clear research gap exists. A new methodology is needed that is both architecturally agnostic—universally applicable to CNNs and Transformers alike—and capable of providing the deep, class-specific insights that high-fidelity methods promise. The SCAN methodology we propose aims to resolve these specific issues by utilizing encoded representations in a novel, reconstruction-based framework.

\section{Methodology}
The core objective of SCAN is to generate a visual explanation that identifies salient regions and reconstructs the specific visual features(e.g., shape, color, texture) that a target model utilizes for its predictions. To achieve this, we frame the problem as learning a compressed representation of class-discriminative features, guided by the principle of reconstruction fidelity. Our methodology comprises three main stages: (1) formulating a saliency-guided input by creating a disparity in feature information, (2) designing a learning objective based on the IB principle to identify and reconstruct information-rich regions, and (3) implementing a decoder network to realize this objective.



\subsection{Concepts}

\begin{figure*}[!h]
\centering
\includegraphics[width=1\linewidth]{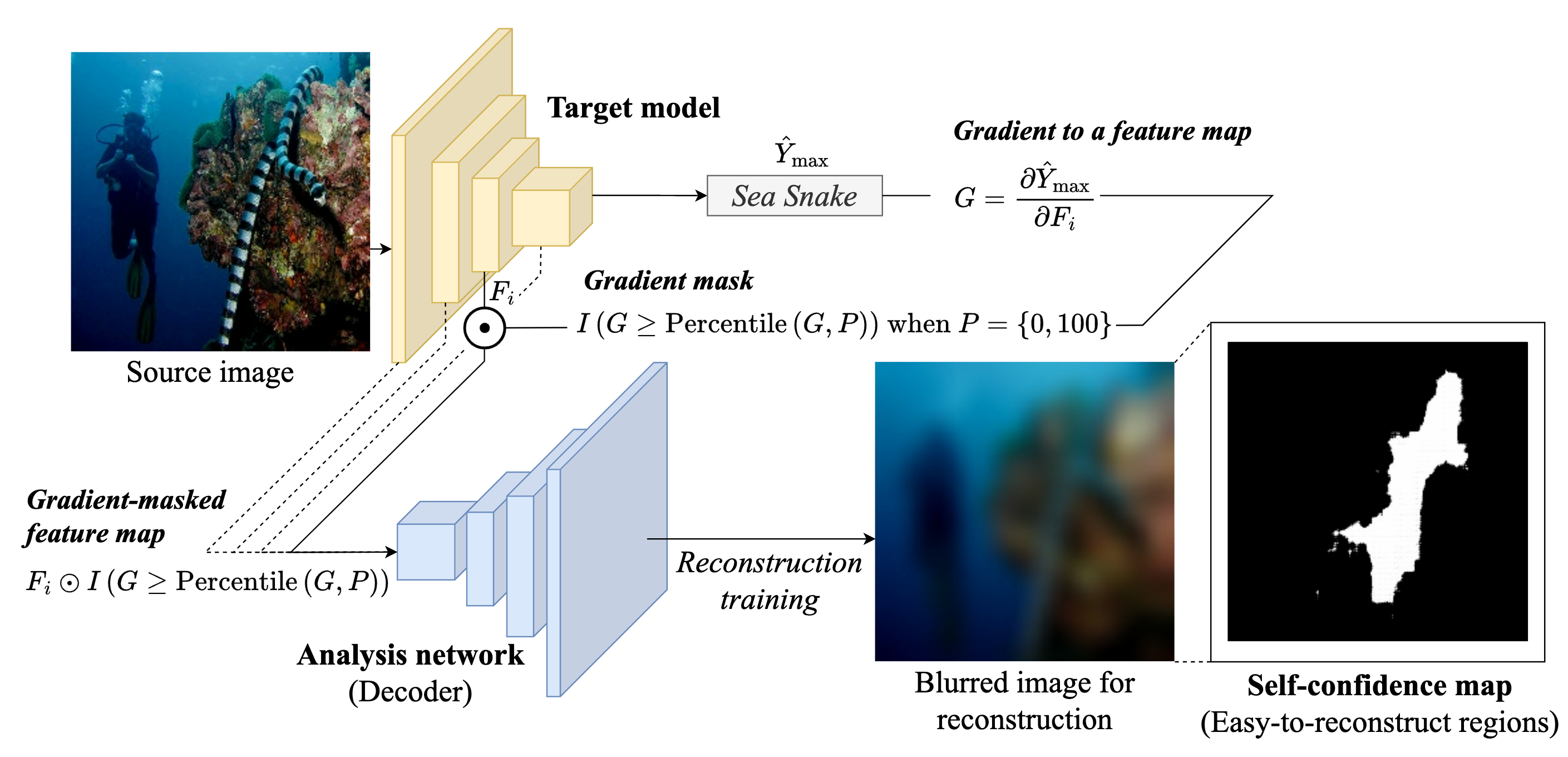}
    \caption{SCAN process. Feature maps are extracted from the target model and reconstructed, and then important regions containing significant information are visualized using the self-confidence map.
    }
    \label{fig1}
\end{figure*}

As Figure \ref{fig1} illustrates, the SCAN framework operates as follows. First, we extract encoded representations(feature maps) $F$ from an intermediate layer of a pre-trained target model. Concurrently, we compute a gradient map $G$ for a specific class, which quantifies the importance of each feature for prediction. This masked representation $X$ is then fed into the SCAN Decoder Network. The decoder is trained with a dual objective: to reconstruct the original input image $Y$ and simultaneously generate a Self-Confidence Map $\hat{C}$ based on \textbf{IB theory}. This map highlights the most informative and easily reconstructible regions in $X$, thus providing a detailed visual explanation.

\subsection{Gradient-masked Feature Map}
To enhance the precision of visual explanations, SCAN applies gradient-masked feature maps. This process generates gradient maps for specific classes and uses these maps as masks to filter semantic features. This filtering ensures that only features strongly linked to specific classes are retained, resulting in more precise visual explanations.

To generate a gradient-masked feature map, we begin with a feature map and a gradient map. The key idea is to filter the feature map using information from the gradient map, highlighting the most relevant features for the model’s decision-making process.

To further refine feature selection, we apply a percentile threshold to the gradient map, considering only the top $P$\% of gradient values and filtering out less relevant features. This is expressed in Equation \ref{eq10}. Here, $F$ represents the feature map, and $G$ represents the gradient map.

\begin{equation}
    \hat{F}=F \odot I\left( G \geq \text{Percentile}(G,P) \right)
    \label{eq10}
\end{equation}

\begin{itemize}
    \item $\text{Percentile}(G,P)$ calculates the $P$ percentile of gradient map $G$. This value serves as the threshold. 
    We randomly set the value of $P$ between 70\% and 100\% during training, and set it to 95\% during the inference.
    \item $I(G\geq\text{Percentile}(G,P))$ creates a binary mask that retains only the gradient values above the percentile $P$.
\end{itemize}

Applying this mask to feature map $F$, produces a gradient-masked feature map that emphasizes the most critical regions for the model’s decision-making process, ensuring that only features strongly associated with specific classes are included for more detailed and precise visual explanations.

\subsection{Information Bottleneck Theory}

The IB theory is a fundamental concept in deep learning, widely applied to identify relationships and patterns within high-dimensional data\cite{ibt1,ibt2,ibt3,ibt4}.

As expressed in Equation \ref{eq:ibt}, this theory introduces a compressed space $T$ between the input data $X$ and target output $Y$. Space $T$ is designed to retain only the information necessary for predicting $Y$, while discarding irrelevant details, thereby maximizing predictive efficiency.
\begin{equation}
    \min_{P(T \mid X)} \left[ I(X;T) - \beta I(T;Y) \right]
    \label{eq:ibt}
\end{equation}

\subsubsection{Information Bottleneck Theory of SCAN}
SCAN leverages the IB theory to visualize salient information within semantic features. Specifically, SCAN treats semantic features extracted from specific neural network layers as the input data X and designates the original input image as the target output Y. This framework enables the model to learn reconstruction mappings from semantic features back to original images, similarly to autoencoder architectures.

Within this framework, the compressed space T is designed to represent ``easy-to-reconstruct regions,'' which include peripheral information from semantic features that significantly influence the model’s decision process.

The implementation uses dual loss functions to support the autonomous learning of self-confidence maps. The analysis network architecture outputs a self-confidence map as a fourth channel alongside the reconstructed image. The confidence loss(Equations \ref{eq1}-\ref{eq4}) constrains the self-confidence map to the area specified in Equation \ref{eq3}, while the reconstruction loss(Equation \ref{eq5}) expands self-confidence map regions by increasing reconstruction penalties through $\alpha$-scaling when confidence approaches unity.

This interaction between loss functions serves two purposes: Confidence loss shapes the compressed space T, and reconstruction loss assigns opportunity costs to ``easy-to-reconstruct regions''. Consequently, the model identifies pixels that are easy to reconstruct within constrained self-confidence map regions. This mechanism enforces prioritization in pixel selection, driving the latent space T toward high-efficiency reconstruction information and producing the IB effect. As shown in Figure \ref{fig1}, this methodology enables the visualization of critical regions and reconstructed information from semantic features.

\subsection{Loss functions with Information Bottleneck}

\subsubsection{Confidence Loss}
Equation \ref{eq0} represents the division of the model’s four-channel output into the reconstructed image and the self-confidence map. Specifically, $\hat{Y}_r$ corresponds to the reconstructed RGB image, while $\hat{Y}_c$ denotes the self-confidence map.
\begin{equation}
    \hat{Y}_r = \hat{Y}_{\{1,2,3\}} ,
    \quad \hat{Y}_c = \hat{Y}_4
    \label{eq0}
\end{equation}

The self-confidence map $\hat{C}$ is defined to range between 0 and 1, as shown in Equation \ref{eq1}. A stretching sine function is applied as the activation function to express the reconstruction confidence of each pixel. This function limits output values to a bounded range, preventing infinite growth, avoiding gradient vanishing issues, and ensuring balanced reconstruction results.
\begin{equation}
    \hat{C}_i=\frac{\frac{\hat{Y}_c}{|\hat{Y}_c|} \sin \left( \frac{2\pi|\hat{Y}_c|}{8+0.15|\hat{Y}_c|} \right) +1}{2} ,
    \quad \hat{C}_\mu = \text{mean}\left(\hat{C}\right)
    \label{eq1}
\end{equation}

Equation \ref{eq3} specifies the target area size of the self-confidence map, which is controlled by the hyperparameter $\alpha$. According to Equation \ref{eq2}, as $\omega$ approaches $A_c$, it converges towards 0. Applying $\omega$ from Equation \ref{eq4} ensures that the self-confidence map area $\hat{C}_\mu$ converges to $A_c$. This enforces the maintenance of a limited compression space $T$.
\begin{equation}
    A_c=\frac{1}{1+\alpha}
    \label{eq3}
\end{equation}

\begin{equation}
    \omega=\frac{\left( \hat{C}_\mu-A_c \right)^2}{\hat{C}_\mu \left( 1-\hat{C}_\mu \right)} 
    \label{eq2}
\end{equation}


To address the information loss that occurs when downsampling the original image to the feature map's spatial dimensions, we use a blurred version of the original image, denoted as $\tilde{Y}$, as the reconstruction target. This is necessary because high-frequency details lost during downsampling are impossible to restore from the lower-resolution feature map. We generate $\tilde{Y}$ by applying a Gaussian blur to the original image, as shown in Equation \ref{eq3.5}.

The kernel size $k$ and standard deviation $\sigma$ for the Gaussian blur are dynamically set based on the image and feature map sizes to ensure an appropriate level of smoothing. Given an original image size of $s_k \times s_k$ and a feature map size of $s_\text{f} \times s_\text{f}$, we define $k$ and $\sigma$ as follows:

\begin{itemize}
    \item Kernel Size ($k$): $k=2\lfloor \frac{s_k}{s_\text{f}} \rfloor+1$
    \item Standard Deviation ($\sigma$): $\sigma=\frac{s_k}{2s_\text{f}}$
\end{itemize}

\begin{equation}
    \tilde{Y}=\text{GaussianBlur}\left( Y, k, \sigma\right)
    \label{eq3.5}
\end{equation}

The constant $\lambda$ determines the flexibility of allowable errors as the self-confidence map area
$\hat{C}_\mu$ converges. A higher $\lambda$ reduces this flexibility, forcing the model to adhere more strictly to the target area. In SCAN, the default value of $\lambda$ is set to 0.1, providing a balance between flexibility and precision during training.

\begin{equation}
    \text{Loss}_c=\left( 1+\omega \right) \left( \left| \left| \tilde{Y}-\hat{Y}_r \right| \right|_2^2 + \lambda \right) - \lambda
    \label{eq4}
\end{equation}

\subsubsection{Reconstruction Loss}
The reconstruction Loss, defined in Equation \ref{eq5}, is an MSE loss that increases as the self-confidence value increases. The $\alpha\hat{C}_i$ term increases the loss in high-confidence areas, encouraging the model to prioritize activating regions with low loss as self-confidence areas.

\begin{equation}
    \text{Loss}_{r}=\frac{1}{N} \sum_{i=1}^N  
    \alpha\hat{C}_{i} \left( \tilde{Y}_{i}-\hat{Y}_{ri} \right)^2 + \left( 1-\hat{C}_{i} \right) \left( \tilde{Y}_{i}-\hat{Y}_{ri} \right)^2
    \label{eq5}
\end{equation}


Consequently, since the area of the self-confidence map is constrained by Equation \ref{eq4}, the model outputs self-confidence map regions containing the most important information within the semantic features under the influence of the IB Theory. The overall loss function is represented by Equation \ref{eq:overall}.

\begin{equation}
    \text{Loss} = \text{Loss}_c + \text{Loss}_r
    \label{eq:overall}
\end{equation}

\subsection{Analysis Networks}
We designed an analysis network (decoder) based on transformer\cite{23,18} and ResNet\cite{21,22} architectures to visualize and analyze representations. When analyzing the CNN architecture model, it is structured to increase to the size of the original image by repeatedly applying 2 Residual Modules and 1 Transposed Conv multiple times. When analyzing the Transformer model, after passing through 4 Attention Modules, 2 Residual Modules and 1 Transposed Conv are repeatedly applied multiple times to reach the size of the original image. The decoder outputs four channels: one for the self-confidence map and three for image reconstruction. 

\begin{figure*}[htb!]
    \centering
    \begin{subfigure}{1\linewidth}
        \centering
        \includegraphics[width=0.9\textwidth]{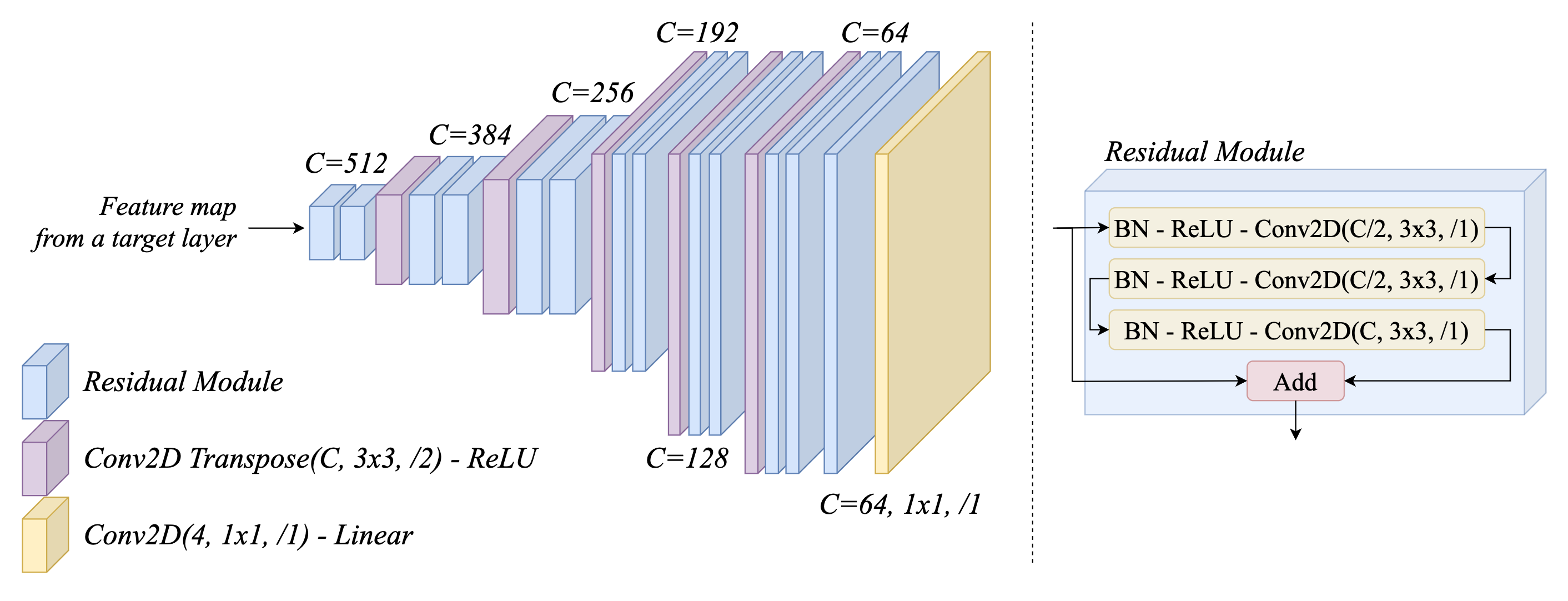}

        \caption{Residual-based analysis network structure for CNN models. It consists only of CNNs, making it easy to reconstruct the encoded representation from the CNN model structure.}
        \label{fig3a}
    \end{subfigure}
    \hfill
    \begin{subfigure}{1\linewidth}
        \centering
        \includegraphics[width=0.9\textwidth]{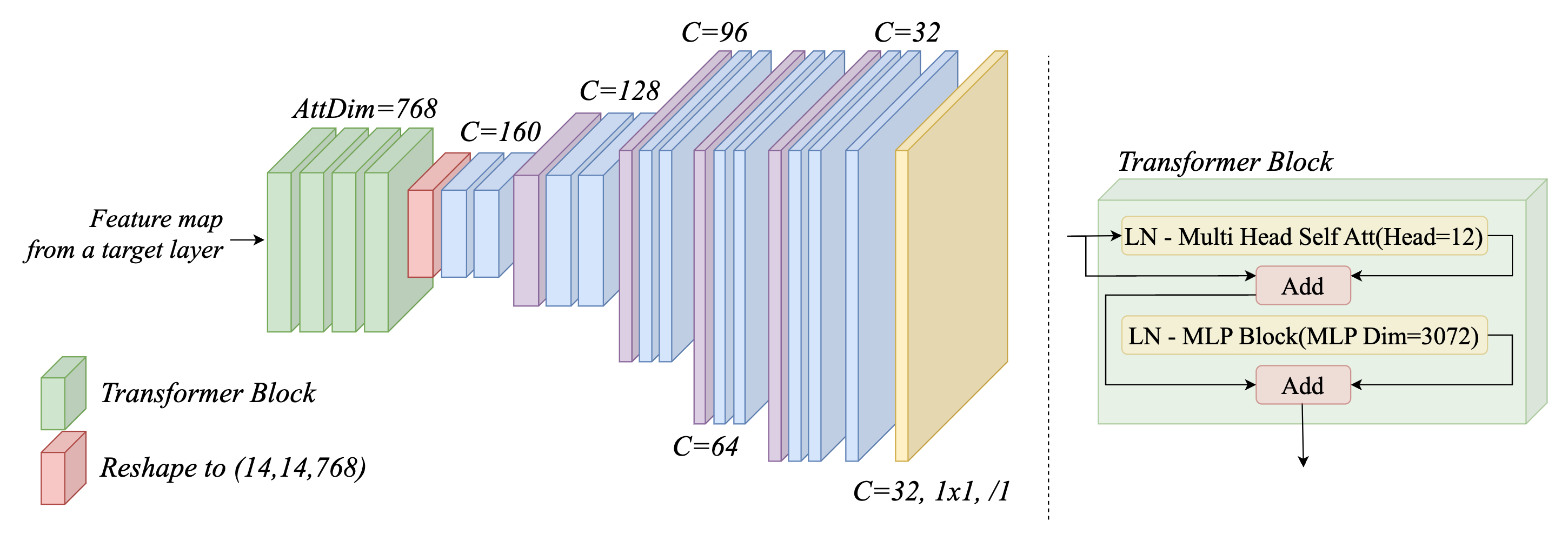}
        \caption{Transformer-based analysis network structure for transformer models. It is organized based on Transformer Blocks, which makes it easy to reconstruct the encoded representation formed from the Transformer model structure.}
        \label{fig3a}
    \end{subfigure}
    \caption{Analysis networks for CNN and transformer models. The ResNet-based decoder is optimized for CNN model structures, while the transformer-based decoder is designed for transformer model structures.}
    \label{fig3}
\end{figure*}

Figure \ref{fig3} illustrates the structure of the Residual-based analysis network used for CNN-based models and the transformer-based analysis network used for transformer-based models. The ResNet-based analysis network can vary in depth depending on the resolution of the input feature map; when using feature maps from shallower layers, the network depth is reduced. In contrast, the transformer-based analysis network always uses the same structure since the resolution of the feature maps remains constant.

\begin{table*}[!h]
\centering
\caption{Quantitative comparison on \textit{ImageNet}, \textit{CUB-200}, and \textit{Food-101}. The symbol {$\dagger$} indicates that the result comes from Expl.\cite{exp}}
\label{tab1_imagenet}


\resizebox{0.75\textwidth}{!}{
\begin{tabular}{l|rrrrrr}
\toprule
\multicolumn{7}{c}{ImageNet}\\
\midrule
\textbf{Method} & \textbf{AUC-D} $\uparrow$ & \textbf{Neg AUC} $\uparrow$ & \textbf{Pos AUC} $\downarrow$ & \textbf{Drop\%} $\downarrow$ & \textbf{Inc\%} $\uparrow$ & \textbf{Win\%} $\uparrow$ \\
\midrule
Raw Attn. & 21.56\%{$^\dagger$} & 45.55\%{$^\dagger$} & 23.99\%{$^\dagger$} & 68.26\% & 26.18\% & 9.31\% \\
Rollout & 33.05\%{$^\dagger$} & 53.1\%{$^\dagger$} & 20.05\%{$^\dagger$} & 53.98\% & 23.42\% & 16.62\% \\
LRP & 1.55\%{$^\dagger$} & 43.49\%{$^\dagger$} & 41.94\%{$^\dagger$} & 40.93\% & 16.14\% & 29.97\% \\
Part. LRP & 30.85\%{$^\dagger$} & 50.49\%{$^\dagger$} &  19.64\%{$^\dagger$} & 84.86\% & 29.42\% & 4.50\% \\
Expl. & 37.13\%{$^\dagger$} & 54.16\%{$^\dagger$} & 17.03\%{$^\dagger$} & 85.87\% & 24.16\% & 3.05\% \\
GradCAM & 7.46\%{$^\dagger$} & 41.52\%{$^\dagger$} & 34.06\%{$^\dagger$} & 72.92\% & 28.93\% & 8.38\% \\
Att. CAM & 38.52\% & 47.24\% & 8.72\% & 45.99\% & 7.77\% & 10.91\% \\
LIME & 32.55\% & 47.56\% & 15.01\% & 38.26\% & 21.62\% & 34.71\% \\
RISE & 28.88\% & 44.16\% & 15.28\% & 40.98\% & 24.66\% & 31.22\% \\
\textbf{SCAN} (ours) & \textbf{36.87\%} & \textbf{49.29\%} & \textbf{12.42\%} & \textbf{65.33\%} & \textbf{27.15\%} & \textbf{11.22\%} \\
\midrule
\midrule
\multicolumn{7}{c}{Caltech-UCSD Birds-200-2011}\\
\midrule
Raw Attn. & 14.99\% & 48.91\% & 33.92\% & 20.46\% & 4.60\% & 14.34\% \\
Rollout & 77.54\% & 81.28\% & 3.74\% & 11.67\% & 3.89\% & 24.91\% \\
LRP & 1.18\% & 43.07\% & 41.89\% & 22.41\% & 1.88\% & 30.63\% \\
Part. LRP & 34.71\% & 60.47\% & 25.76\% & 40.44\% & 5.97\% & 5.61\% \\
Expl. & 76.17\% & 81.90\% & 5.73\% & 68.81\% & 7.00\% & 2.11\% \\
GradCAM & 35.39\% & 59.67\% & 24.29\% & 26.51\% & 5.79\% & 12.13\% \\
Att. CAM & 75.24\% & 81.09\% & 5.85\% & 19.98\% & 4.70\% & 19.39\% \\
LIME & 64.89\% & 77.48\% & 12.59\% & 4.58\% & 2.98\% & 45.04\% \\
RISE & 47.80\% & 72.72\% & 24.92\% & 8.62\% & 4.37\% & 38.69\% \\
\textbf{SCAN} (ours) & \textbf{77.80\%} & \textbf{83.53\%} & \textbf{5.73\%} & \textbf{11.36\%} & \textbf{4.43\%} & \textbf{31.62\%} \\
\midrule
\midrule
\multicolumn{7}{c}{Food-101}\\
\midrule
Raw Attn. & 29.74\% & 53.12\% & 23.38\% & 45.35\% & 8.62\% & 4.01\% \\
Rollout & 21.37\% & 51.72\% & 30.35\% & 26.26\% & 7.33\% & 12.93\% \\
LRP & 0.35\% & 39.90\% & 39.55\% & 25.44\% & 3.31\% & 26.20\% \\
Part. LRP & 33.50\% & 56.99\% & 23.49\% & 69.71\% & 11.53\% & 0.81\% \\
Expl. & 36.56\% & 61.64\% & 25.08\% & 87.01\% & 9.53\% & 0.12\% \\
GradCAM & 36.50\% & 55.82\% & 19.31\% & 48.09\% & 7.96\% & 3.81\% \\
Att. CAM & 43.57\% & 63.05\% & 19.48\% & 51.52\% & 8.39\% & 4.02\% \\
LIME & 36.61\% & 67.86\% & 31.25\% & 13.90\% & 6.66\% & 30.96\% \\
RISE & 26.39\% & 62.14\% & 35.75\% & 13.70\% & 7.62\% & 28.34\% \\
\textbf{SCAN} (ours) & \textbf{37.29\%} & \textbf{63.43\%} & \textbf{26.14\%} & \textbf{58.11\%} & \textbf{10.60\%} & \textbf{3.05\%} \\
\bottomrule
\end{tabular}
} 
\end{table*}

\section{Experiments}
\subsection{Evaluation metrics}
To evaluate the methods’ performance, we measured prediction accuracy and visualization quality using the three widely used metrics(Drop\%, Increase\%, Win\%)\cite{rise}, as well as Positive AUC and Negative AUC. The AUC metrics utilize sequential perturbation to enhance reliability\cite{exp}.

\textbf{Drop percentage} measures performance loss when images are masked by a confidence map, with lower values reflecting stronger feature retention. 

\textbf{Increase percentage} quantifies performance gains when masked images are used, with higher values indicating more effective feature highlighting. 

\textbf{Win percentage} calculates the proportion of masked images that outperform the original, signifying overall improvement.

\textbf{Positive AUC} measures how quickly the model's accuracy drops when the most important pixels are removed first. A sharp drop means that the explanation correctly identified key features.

\textbf{Negative AUC} measures how well the model's accuracy is maintained when the least important pixels are removed first. Good performance indicates that the explanation correctly identified irrelevant features.

SCAN and LIME generate more precise boundaries compared to other methods. In contrast, most existing approaches such as Explainability\cite{exp} and GradCAM create a 14$\times$14 or 7$\times$7 saliency map and then upsample it to 224$\times$224. Therefore, for a fair comparison, the saliency maps from SCAN and LIME were first reduced to 14$\times$14 or 7$\times$7, then upsampled back to 224$\times$224 before measuring the metrics. Specifically, the maps were reduced to 14$\times$14 for Transformer models and to 7$\times$7 for CNN models.

\subsubsection{A Reliable Metric for Comprehensive Evaluation}

Chefer et al. \cite{exp} have indicated that metrics such as Drop\%, Increase\%, and Win\% exhibit low reliability. To enhance the evaluation reliability, we conducted an experiment to assess the response of these metrics to meaningless explanations. For each data instance, we generated ten random saliency maps and measured each metric.

The results revealed that, for these random maps, the Drop\% was 45.4\%, Increase\% was 17.38\%, and Win\% was 20.39\%. These findings suggest that saliency maps from visual explanation methods can be misinterpreted as effective even when they are meaningless. Furthermore, the significant differences in the scales of these metrics make direct comparison challenging.

In contrast, Pos AUC and Neg AUC were measured at 14.15\% and 14.16\%, respectively, indicating that both metrics operate on an identical scale. This implies that using the difference, \textbf{$\text{Neg AUC} - \text{Pos AUC}$}, can accurately quantify the explanatory power of a saliency map, as its value converges to zero for meaningless explanations. Therefore, we define and employ an additional single metric, the AUC Difference(\textbf{AUC-D}), for a comprehensive evaluation.

\subsection{Dataset and Training}
The model was trained for 5 epochs on the ImageNet, 50 epochs on the Caltech-UCSD Birds-200-2011(CUB), and 10 epochs on the Food-101 datasets. The experiments were conducted on a server equipped with an AMD Ryzen ThreadRipper PRO 3955WX CPU, 192 GB RAM, and four Nvidia A5000 GPUs.

Across all experiments, the hyperparameters were set to $\alpha=4$ and $P=95\%$. For applying SCAN, feature maps were extracted from the 6th attention layer for transformers, and from the final convolutional layer for CNNs. 

\subsubsection{Augmentation with Gradient masking }
Traditional computer vision augmentation techniques, such as Color Jitter, Cutout\cite{24}, CutMix\cite{25}, and Mixup\cite{26}, are ineffective for SCAN because they do not consider the semantic meaning of feature maps. To address this limitation, SCAN applies a specialized augmentation method using gradient-masked feature maps. This approach performs percentile filtering on the gradient map, randomly retaining between 70--100\% of the top gradients, enabling focused learning on the critical representations.

\subsection{Experimental Results}
\subsubsection{Comparison of visual explanations for transformer across datasets}
We conducted a comprehensive quantitative evaluation of SCAN against several state-of-the-art visual explanation methods on the ImageNet, CUB, and Food-101 datasets. All evaluations were performed on the ViT-b16. Although Drop\%, Inc\%, and Win\% are widely used in prior work, they may exhibit sensitivity to confidence scaling and threshold effects~\cite{exp}. Therefore, in this study, we adopt AUC-based metrics as the primary evaluation criteria, while reporting Drop\%, Inc\%, and Win\% as complementary indicators for reference. The results are presented in Table \ref{tab1_imagenet}. As shown, SCAN demonstrated highly competitive or superior performance across most standard evaluation metrics.


\begin{figure*}[!h] 
\centering
\includegraphics[width=0.98\textwidth]{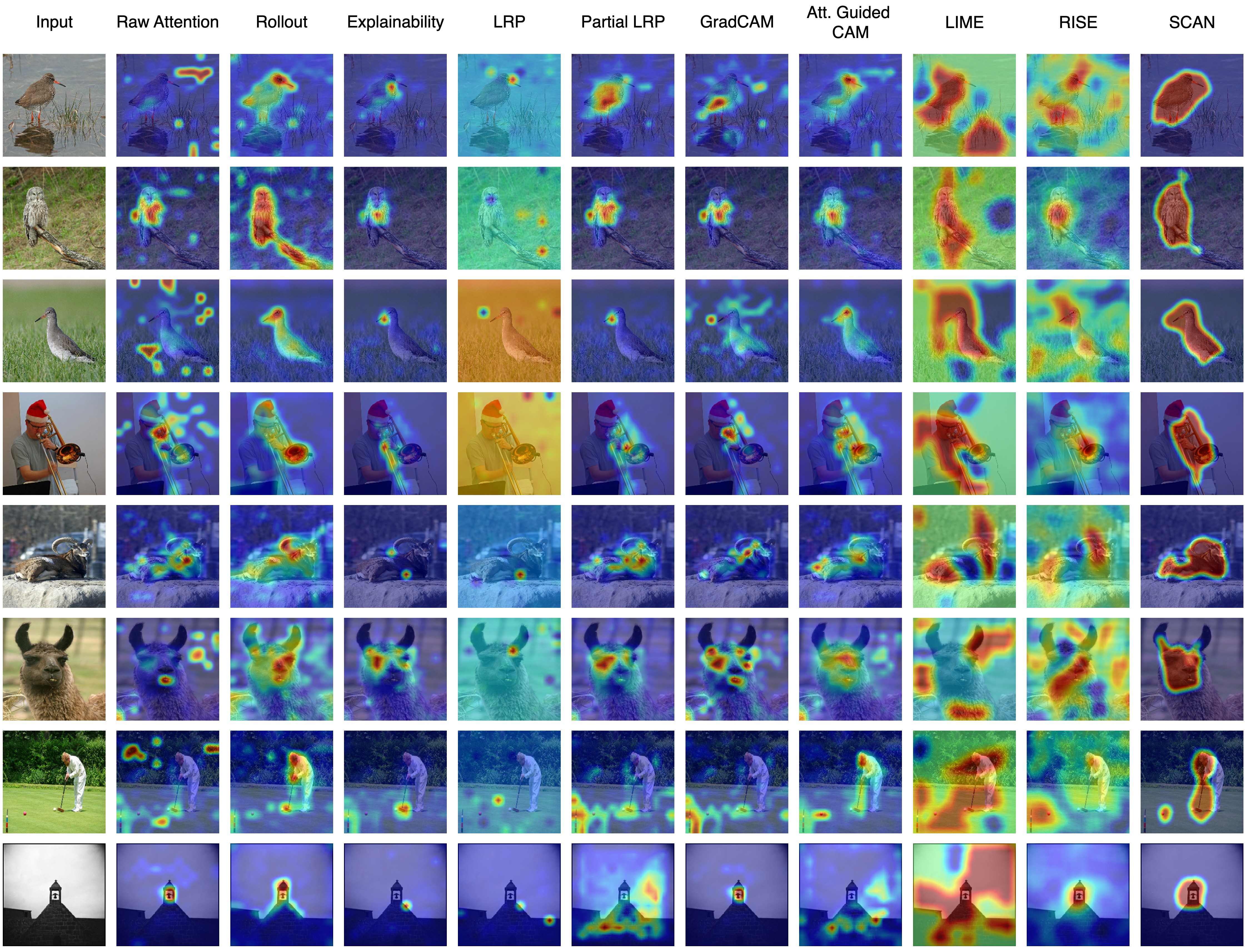}
    \caption{Qualitative comparison of visual explanation methods for a ViT-b16 model trained on ImageNet. Compared to baselines such as Raw Attention, Rollout, and others, SCAN generates a more coherent and object-focused explanation. }
    \label{fig5}
\end{figure*}

Quantitative comparisons on the ImageNet dataset are presented in Table~\ref{tab1_imagenet}. SCAN achieved an AUC-D of $36.87\%$, demonstrating competitive performance comparable to Explainability($37.13\%$). Notably, SCAN achieved a low Positive AUC of $12.42\%$ among all methods, indicating superior capability in identifying features crucial for the model's decision-making. Furthermore, when compared to Explainability, which showed the most similar AUC-D performance, SCAN exhibited significant improvements in faithfulness metrics. Specifically, SCAN's Drop\% was $65.33\%$, which is $20.54\text{p.p.}$ lower than that of Explainability. Similarly, its Inc\% and Win\% were higher by $2.99\text{p.p.}$ and $8.17\text{p.p.}$, respectively. Consequently, these results confirm that the saliency maps generated by SCAN effectively capture regions that are highly critical for prediction.

These compelling results are consistently observed on the CUB and Food-101 datasets as well. SCAN achieved high AUC-D scores on these diverse datasets, while maintaining competitive performance across various other metrics.

In conclusion, although SCAN does not rank first in all metrics, it particularly excels at identifying the most and least critical features for the model's prediction. This indicates that SCAN provides an overall superior and well-balanced visual explanation.

\subsubsection{Comparison of visual explanations for CNNs}

Table \ref{tab3} presents the quantitative evaluation results for the CNN models. SCAN achieves the best AUC-D of 37.29\%, demonstrating its outstanding ability to accurately identify crucial features. This result is particularly noteworthy given that the exact same method is applied to both Transformer and CNN architectures. 


\begin{table}[!h]
  \caption{Performance comparison of SCAN and other methods on the ImageNet dataset, evaluated on the ResNet50V2. For the clear comparison, we selected several prominent recent methods as baselines. }
  \label{tab3}
  \centering
  \begin{tabular}{l|r|rr}
    \toprule
        \textbf{Methods} & \textbf{AUC-D} $\uparrow$ & \textbf{Neg AUC} $\uparrow$ & \textbf{Pos AUC} $\downarrow$  \\
    \midrule
        XGradCAM 
        & 32.72\% & 40.45\% & 7.73\% \\
        GradCAM++
        & 35.42\% & 41.92\% & 6.50\% \\
        LayerCAM
        & 36.27\% & 42.42\% & 6.15\% \\
        LIME
        & 28.65\% & 37.36\% & 8.71\% \\
        RISE
        & 21.37\% & 28.76\% & 7.39\% \\
    \midrule
        \textbf{SCAN} (ours)
        & \textbf{37.29\%} & \textbf{42.87\%} & \textbf{5.58\%}  \\
    \bottomrule
  \end{tabular}
\end{table}

\begin{figure*}[!h]
\centering
\includegraphics[width=0.7\linewidth]{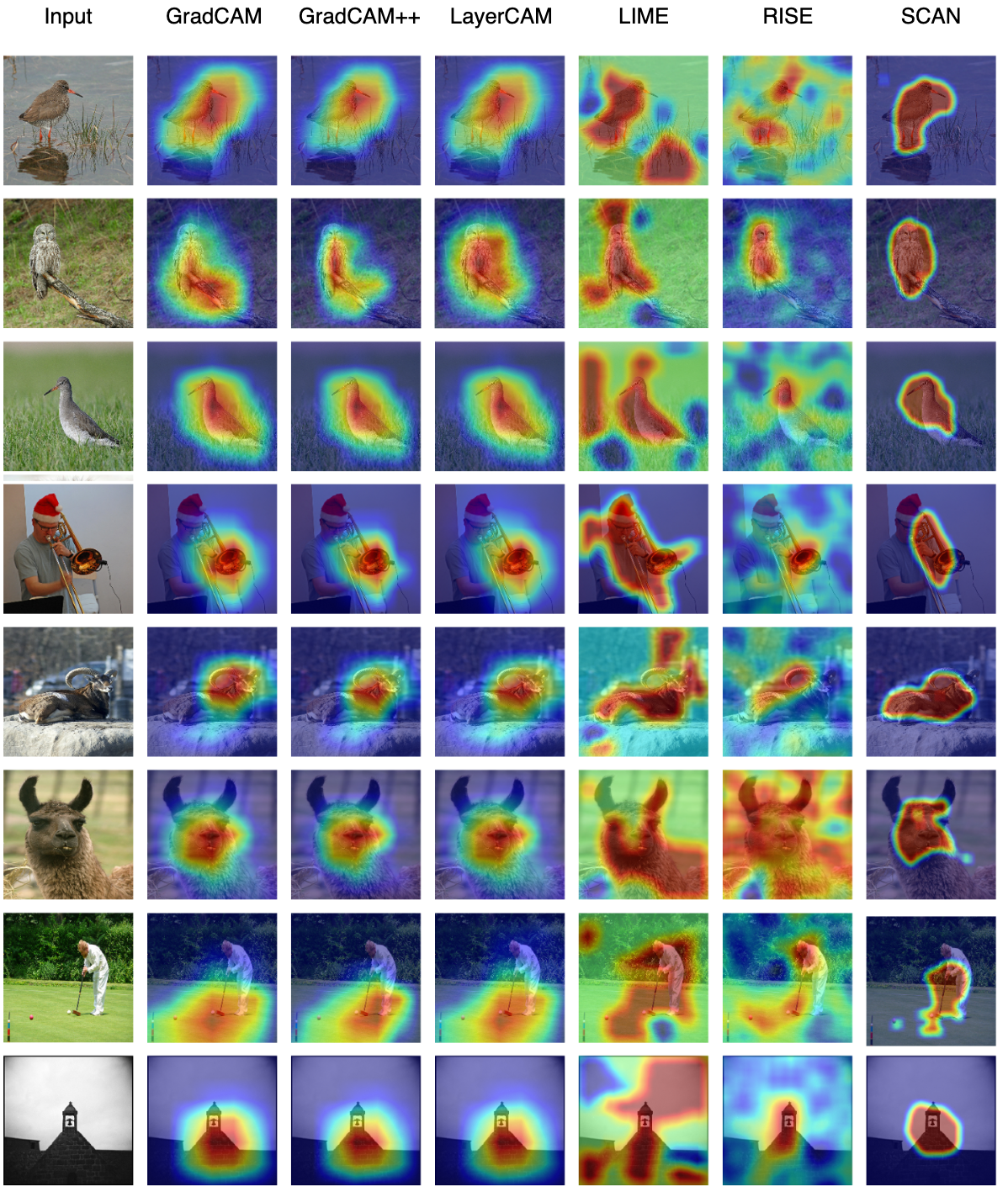}
    \caption{Qualitative comparison of SCAN and other methods on ResNet50V2. While conventional methods generate abstract saliency maps, SCAN produced more distinct explanations with clear object boundaries.}
    \label{fig5.5}
\end{figure*}

\subsubsection{Qualitative Comparison}

Figure \ref{fig5} presents a qualitative comparison of visual explanations for Transformer-based models, evaluating SCAN against existing interpretability techniques. Transformer-specific methods like Raw Attention and Rollout, which are based on the aggregation of attention scores, often produce fragmented saliency maps. Consequently, they tend to indiscriminately highlight irrelevant regions, such as the background, and frequently fail to capture the object's complete form. In contrast, SCAN’s explanations include minimal background noise and precisely localize the region containing the subject. This capability stems from SCAN's ability to integrate information from multiple patch-based representations, enabling it to provide a  holistic and meaningful explanation of the features deemed important by the target model. This proficiency in precisely segmenting the object from the background directly contributes to its exceptional performance on the Negative AUC metric.

Figure \ref{fig5.5} qualitatively compares SCAN with various CAM-based explanation methods on CNN models. While existing techniques such as GradCAM, GradCAM++, and LayerCAM successfully identify the approximate location of the target object, their explanations tend to include significant irrelevant background. Moreover, the blurry boundaries of their saliency maps make precise localization of the key subject challenging. In contrast, SCAN includes minimal background and accurately localizes the region containing the subject.

\subsubsection{Comparison across models}
To evaluate the robustness and generalizability of our method, we compared the performance of SCAN against baseline techniques across a diverse range of architectures(DINO, DeiT, VGG16, and ConvNeXt-s).

As detailed in Table \ref{tab2}, SCAN consistently outperformed the other methods regardless of the underlying model architecture. Across all tested models, SCAN achieved the highest AUC-D score. Furthermore, it generally exhibited high Negative AUC and low Positive AUC values. Notably, SCAN demonstrated exceptionally strong performance on modern architectures such as DINO, DeiT, and ConvNeXt, developed nearly a decade after the older VGG16, highlighting its superior explainability for models with complex structures and training paradigms.

\begin{table}[!h]
  \caption{Comparison with the recent methods across various transformer and CNN models not previously tested on the ImageNet.}
  \label{tab2}
  \centering
  \begin{tabular}{l|l|rrr}
    \toprule
    \textbf{Models} & \textbf{Methods} & \textbf{AUC-D} $\uparrow$ & \textbf{Neg} $\uparrow$ & \textbf{Pos} $\downarrow$ \\
    \midrule
    \multirow{3}{*}{\textit{DINO}}
    & Expl. & 17.44\% & 41.06\% & 23.62\% \\
    & LIME & 28.84\% & 49.98\% & 21.13\% \\
    \cmidrule{2-5}
    & \textbf{SCAN} (ours) & \textbf{40.43\%} & \textbf{60.13\%} & \textbf{19.70\%} \\
    \midrule
    \multirow{3}{*}{\textit{DeiT}}
    & Expl. & 17.60\% & 47.62\% & 30.02\% \\
    & LIME & 15.30\% & 46.80\% & 31.50\% \\
    \cmidrule{2-5}
    & \textbf{SCAN} (ours) & \textbf{32.44\%} & \textbf{61.10\%} & \textbf{28.66\%} \\
    \midrule
    \multirow{3}{*}{\textit{VGG16}}
    & LayerCAM & 29.34\% & 33.59\% & 4.25\% \\
    & LIME & 24.82\% & 38.07\% & 13.25\% \\
    \cmidrule{2-5}
    & \textbf{SCAN} (ours) & \textbf{30.98\%} & \textbf{43.00\%} & \textbf{12.02\%} \\
    \midrule
    \multirow{3}{*}{\textit{ConvNeXt-s}}
    & LayerCAM & 21.72\% & 55.35\% & 33.63\% \\
    & LIME & 22.62\% & 62.64\% & 40.02\% \\
    \cmidrule{2-5}
    & \textbf{SCAN} (ours) & \textbf{39.79\%} & \textbf{69.24\%} & \textbf{29.45\%} \\
    \bottomrule
  \end{tabular}
\end{table}

\begin{figure}[!h]
\centering
\includegraphics[width=0.8\columnwidth]{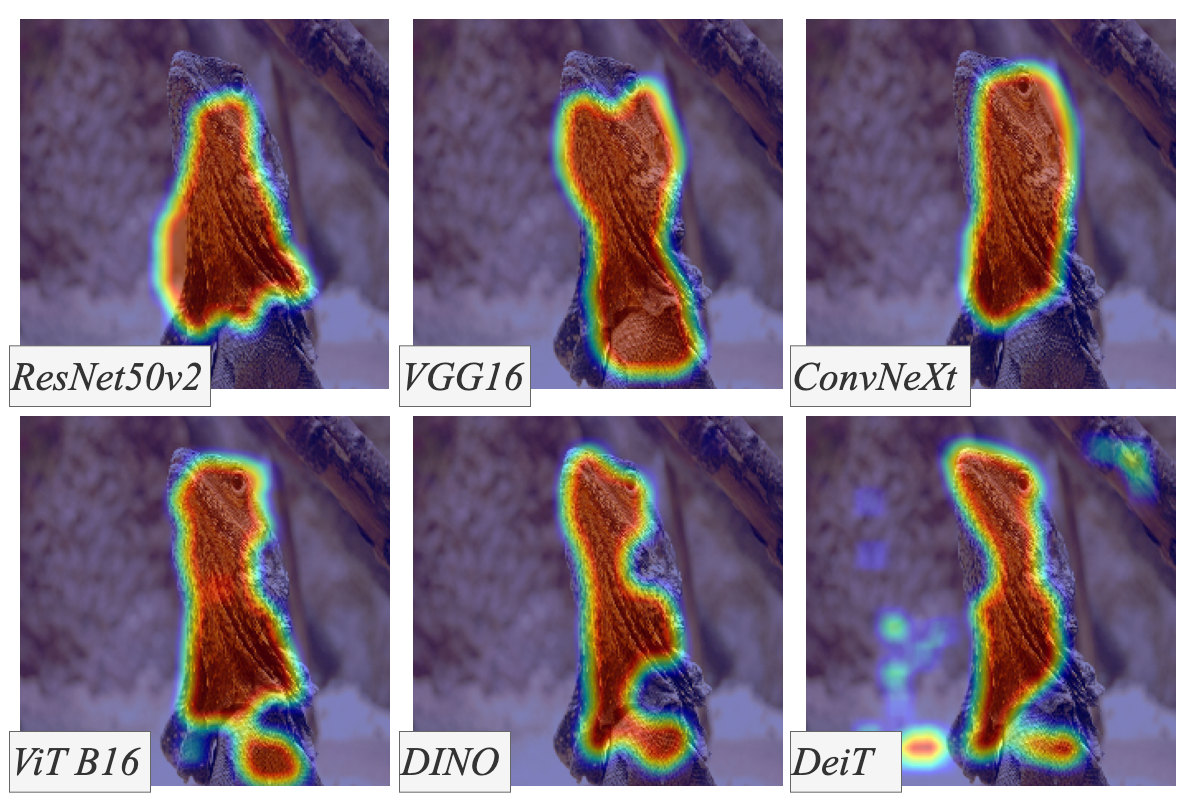}
    \caption{Qualitative results across various models. SCAN consistently generated clear and object-focused explanations.}
    \label{fig6}
\end{figure}

This robustness is also clearly evident in the qualitative evaluation presented in Figure \ref{fig6}. Across all tested architectures, from transformers to CNNs, SCAN consistently generated distinct Self-Confidence Maps that accurately segment the target object. This visual consistency corroborates our quantitative findings, confirming that SCAN operates effectively as an architecture-agnostic explanation method.


\subsubsection{Gradient Mask Percentiles During the Inference}

\begin{figure}[ht]
\centering
\includegraphics[width=0.8\columnwidth]{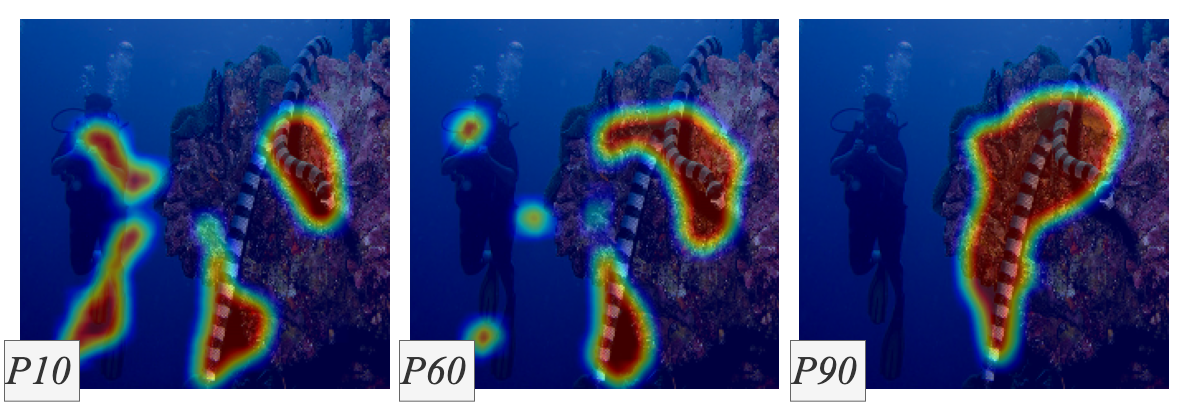}
    \caption{SCAN outputs with different gradient mask percentile values. Explanations were generated using the EfficientNetV2B0.}
    \label{fig9}
\end{figure}

Figure \ref{fig9} presents SCAN outputs with the gradient mask percentiles set to 10\%, 60\%, and 90\%. As the percentile increased, attention shifted toward more critical features, emphasizing only the primary subjects. This reveals that the decision-making process may rely on a combination of the object and its environmental context, a nuanced insight that is often missed by other methods \cite{context1, context2, context4, lime}. Furthermore, the gradient mask percentile can be adjusted during the inference phase, enabling rapid visualization of interpretations at different importance levels.

\subsection{Sanity Check: Verification of Explanations}

Sanity checks are essential to verify whether an explanation method truly reflects the model's learned weights or merely acts as a model-independent edge detector~\cite{adebayo2018sanity}. To ensure this fidelity, we conducted tests using model weights and label randomization. As detailed in Table \ref{tab10}, randomizing the ResNet50V2 weights caused SCAN’s AUC-D score to drop sharply from 37.29\% to 0.01\%. This behavior is consistent with LayerCAM(-0.01\%), confirming that both methods are fundamentally grounded in the features learned. Furthermore, label randomization reduced the score to 31.68\%, demonstrating sensitivity to the network's class-discriminative logic. These results confirm that SCAN's explanations are highly faithful to the model's internal decision process.

\begin{table}[ht]
  \caption{Model Parameter and Label Randomization Test on ImageNet using ResNet50V2. AUC-D scores for LayerCAM and SCAN are compared across weight-randomized, and label-randomized settings to verify fidelity to the target model.}
  \label{tab10}
  \centering
  \begin{tabular}{l|c|cc}
    \toprule
    \multicolumn{4}{c}{\textit{\textbf{Sanity Check}}} \\
    \midrule
    \textbf{Target model} & \textbf{AUC-D} $\uparrow$ & \textbf{Neg AUC} $\uparrow$ & \textbf{Pos AUC} $\downarrow$ \\
    \midrule
    \textbf{LayerCAM} & 36.27\% & 42.42\% & 6.15\% \\
    \quad Random weights & -0.01\% & 0.08\% & 0.09\% \\
    \quad Random label & 32.58\% & 40.03\% & 7.45\% \\
    \midrule
    \textbf{SCAN} & 37.29\% & 42.87\% & 5.58\% \\
    \quad Random weights & 0.01\% & 0.08\% & 0.07\% \\
    \quad Random label & 31.68\% & 38.96\% & 7.28\% \\
    \bottomrule
  \end{tabular}
\end{table}

\subsection{Computation Efficiency}

Table \ref{tab11} compares the average inference time per sample on ImageNet using ResNet50V2, averaged over 64 samples. Gradient-based methods such as XGradCAM, GradCAM++, and LayerCAM are the most efficient, with execution times ranging between 6.66ms and 7.06ms. In contrast, perturbation-based methods like LIME and RISE are significantly slower, requiring 1187.5ms and 11812.5ms respectively due to repeated forward passes. SCAN achieves an inference time of 13.75ms; while slightly higher than gradient-based approaches, it is approximately 86 times faster than LIME and 859 times faster than RISE.

\begin{table}[ht]
  \caption{Comparison of average inference time per sample for SCAN and existing attribution methods on the ImageNet using ResNet50V2.}
  \label{tab11}
  \centering
  \begin{tabular}{l|ccc}
    \toprule
    \multicolumn{4}{c}{\textit{\textbf{Inference Time}}} \\
    \midrule
    \textbf{Methods} & XGradCAM & GradCAM++ & LayerCAM\\
    \midrule
    \textbf{Inference Time} & 6.81ms & 7.06ms & 6.66ms \\
    \midrule
    
    \toprule
    \textbf{Methods} & LIME & RISE & \textbf{SCAN}\\
    \midrule
    \textbf{Inference Time} & 1187.5ms & 11812.5ms & \textbf{13.75ms} \\
    \bottomrule
  \end{tabular}
\end{table}

\section{Ablation Studies for Hyperparameters} \label{A2}
\subsection{Ablation Study for $\alpha$ selection}
We conduct an ablation study to determine the optimal value for the hyperparameter $\alpha$, which controls the scaling of the analysis mask. We evaluated the performance by varying $\alpha$ among {2, 4, 8, 16} for both ViT-b16, a transformer-based model, and ResNet50V2, a CNN-based model.

According to the quantitative results presented in Table \ref{tab7}, $\alpha=4$ achieved the best performance across all metrics, including our primary metric AUC-D, for both models.

\begin{figure}[h]
\centering
\includegraphics[width=0.94\linewidth]{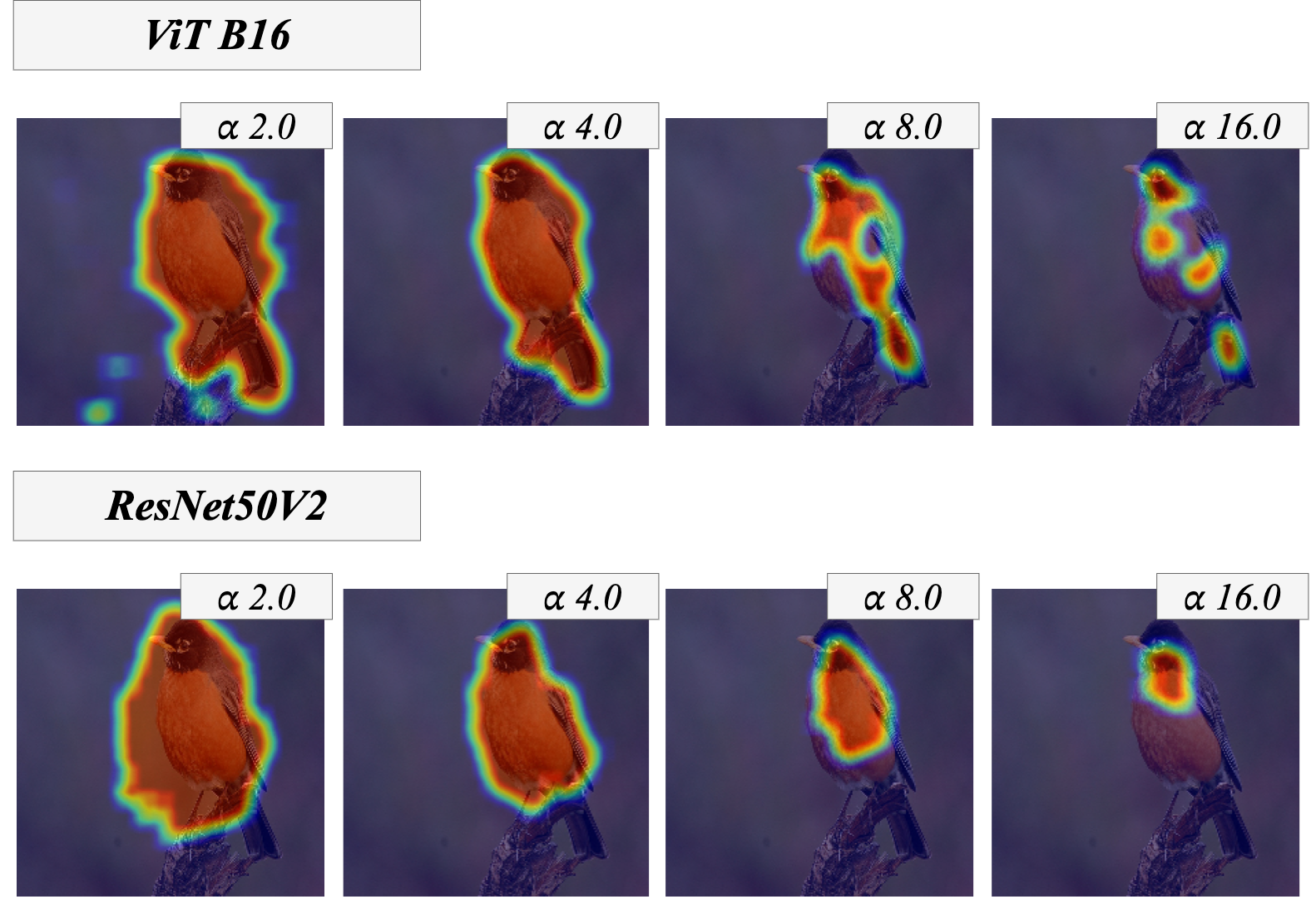}
    \caption{Qualitative ablation study for the hyperparameter $\alpha$. For both ViT-b16 and ResNet50V2, we visualize saliency maps for $\alpha \in \{2, 4, 8, 16\}$. An $\alpha$ of 2 includes excessive background, while values of 8 and 16 are too restrictive and fail to cover the entire object. $\alpha=4$ provides the best balance by accurately highlighting the object of interest.}
    \label{figA3}
\end{figure}

The qualitative results in Figure \ref{figA3} visually substantiate this quantitative analysis. When $\alpha=2$, the saliency map tends to be overly broad, encompassing excessive background context beyond the object. Conversely, increasing $\alpha$ to 8 and 16 causes the map to become too restrictive, failing to cover the entire object. In contrast, $\alpha=4$ provides the best trade-off, accurately localizing the complete object while effectively suppressing background noise.

Therefore, based on this comprehensive quantitative and qualitative analysis, we set $\alpha$ to 4 for all experiments throughout this paper.

\begin{table}[!h]
  \caption{Ablation study on $\alpha$ for transformer(ViT) and CNN(ResNet). The $\alpha=4$ is the best between the four $\alpha$. The percentile $P$ is set as 95. The best-performing configuration is highlighted in bold.}
  \label{tab7}
  \centering
  \resizebox{0.9\linewidth}{!}{
  \begin{tabular}{l|cccc}
    \toprule
    \multicolumn{5}{c}{\textit{ViT-b16}} \\
    \bottomrule
    \multirow{2}{*}{\textbf{Metrics}} & \multicolumn{4}{c}{\textbf{6th Att. layer}}  \\
    \cmidrule{2-3} \cmidrule{4-5}
    & \textbf{$\alpha=2$} & \textbf{$\alpha=4$} & \textbf{$\alpha=8$} & \textbf{$\alpha=16$} \\
    \midrule
    \textbf{AUC-D} $\uparrow$ & 35.76\% & \textbf{36.87\%} & 34.30\% & 14.00\%  \\
    \cmidrule{2-5}
    Neg AUC $\uparrow$ & 48.44\% & \textbf{49.29\%} & 47.56\% & 46.55\%  \\
    Pos AUC $\downarrow$ & 12.68\% & \textbf{12.42\%} & 13.26\% & 32.55\%  \\
    \midrule
    
    \toprule
    \multicolumn{5}{c}{\textit{ResNet50V2}} \\
    \bottomrule
    \multirow{2}{*}{\textbf{Metrics}} & \multicolumn{4}{c}{\textbf{The last conv layer}}  \\
    \cmidrule{2-3} \cmidrule{4-5}
    & \textbf{$\alpha=2$} & \textbf{$\alpha=4$} & \textbf{$\alpha=8$} & \textbf{$\alpha=16$} \\
    \midrule
    \textbf{AUC-D} $\uparrow$ & 35.26\% & \textbf{37.29\%} & 35.82\% & 31.20\%  \\
    \cmidrule{2-5}
    Neg AUC $\uparrow$ & 41.4\% & \textbf{42.87\%} & 42.15\% & 39.08\%  \\
    Pos AUC $\downarrow$ & 6.14\% & \textbf{5.58\%} & 6.33\% & 7.88\%  \\
    \bottomrule
    \end{tabular}
    }
\end{table}

\subsection{Ablation Study for a layer selection}
As summarized in Table \ref{tab6}, we conducted a systematic ablation study to identify the optimal layer depth and scaling factor $\alpha$ for each backbone architecture. For the ViT-B16 model, peak performance was achieved when extracting attention maps from the 6th transformer layer with $\alpha=4$. Analogously, the ResNet50V2 model attained its best results when the final convolutional layer was selected in conjunction with $\alpha=4$.

\begin{figure}[!h]
\centering
\includegraphics[width=0.94\linewidth]{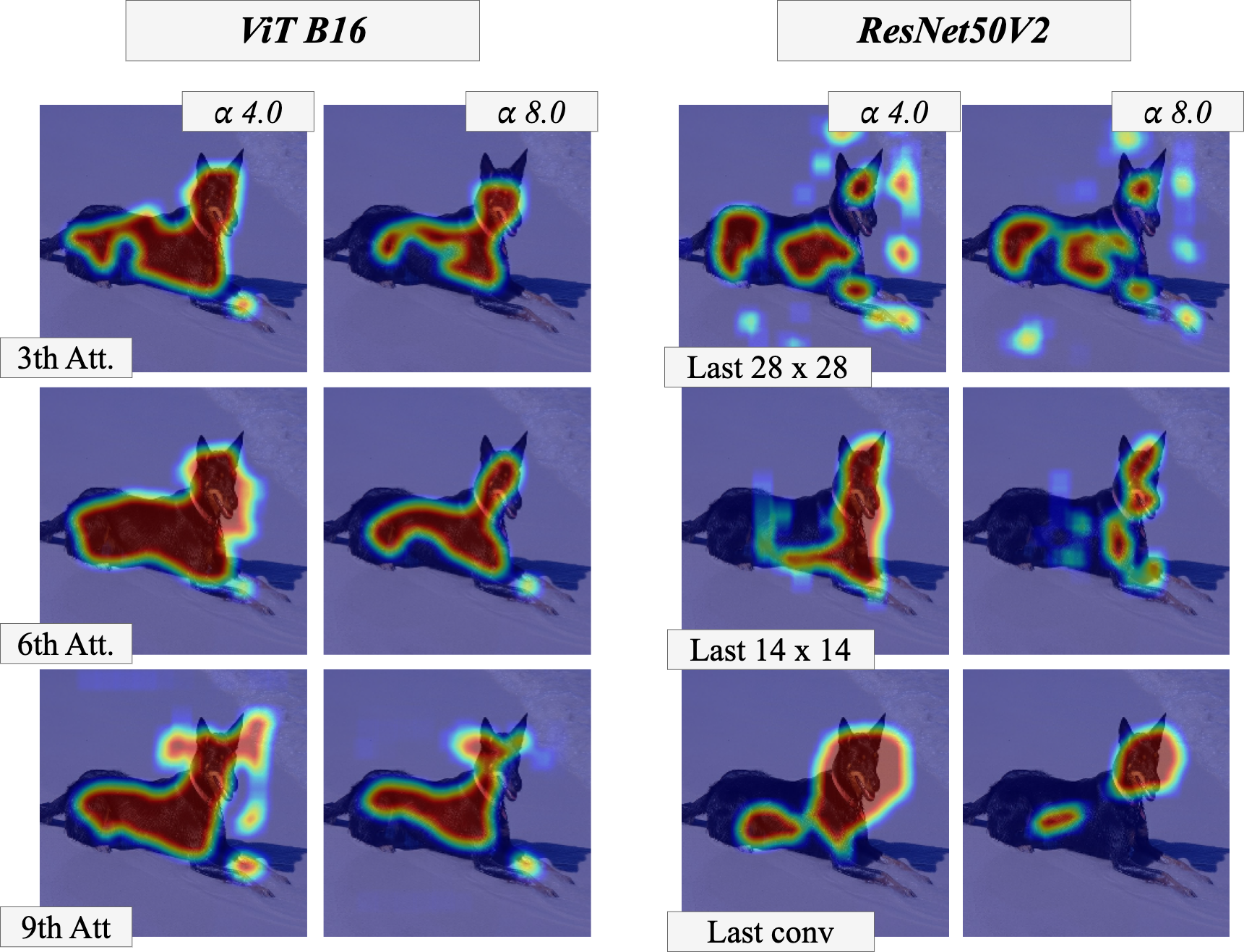}
    \caption{Qualitative ablation study on layer selection for ViT-B16 and ResNet50V2. For ViT, the 6th attention layer provides the best trade-off between diffuse low-level features(3rd layer) and overly sparse high-level features(9th layer). For ResNet, the final convolutional layer produces the most coherent and well-localized saliency map compared to shallower layers.}
    \label{figA2}
\end{figure}

The qualitative results presented in Figure \ref{figA2} corroborate these quantitative findings and provide intuitive insight into the underlying mechanism. For ViT-B16, attention maps derived from shallower layers(e.g., the 3rd attention layer) exhibit diffuse activations that lack spatial precision. Conversely, maps from deeper layers(e.g., the 9th attention layer) become overly sparse, attending exclusively to narrow discriminative regions while neglecting holistic object coverage. The intermediate 6th layer achieves an effective balance, producing saliency maps that accurately delineate the entire object extent with minimal background noise. A consistent pattern emerges for ResNet50V2: earlier convolutional layers generate spatially incoherent and noisy activations, whereas the final convolutional layer yields the most semantically coherent explanations with precise object localization. These observations suggest that intermediate-to-late layers encode representations at an appropriate level of abstraction for generating faithful visual explanations.

\begin{table*}[!h]
  \caption{Ablation study on key hyperparameters for transformer(ViT) and CNN(ResNet). We vary the layer index from which feature maps are extracted and the $\alpha$. The percentile $P$ is set as 95. The best-performing configuration is highlighted in bold.}
  \label{tab6}
  \centering
  \begin{tabular}{l|l|cc|cc|cc}
    \toprule
    \multirow{2}{*}{\textbf{Models}} & \multirow{2}{*}{\textbf{Metrics}} & \multicolumn{2}{c|}{\textbf{9th Att. layer}} & \multicolumn{2}{c|}{\textbf{6th Att. layer}} & \multicolumn{2}{c}{\textbf{3rd Att. layer}} \\
    \cmidrule{3-4} \cmidrule{5-6} \cmidrule{7-8}
    & & \textbf{$\alpha=4$} & \textbf{$\alpha=8$} & \textbf{$\alpha=4$} & \textbf{$\alpha=8$} & \textbf{$\alpha=4$} & \textbf{$\alpha=8$} \\
    \midrule
    \multirow{3}{*}{\textit{ViT-b16}} 
    & \textbf{AUC-D} $\uparrow$ & 34.56\% & 32.67\% & \textbf{36.87\%} & 34.30\% & 33.10\% & 30.84\% \\
    \cmidrule{2-8}
    &  Neg AUC $\uparrow$ & 47.39\% & 46.40\% & \textbf{49.29\%} & 47.56\% & 47.17\% & 45.72\% \\
    &  Pos AUC $\downarrow$ & 12.83\% & 13.73\% & \textbf{12.42\%} & 13.26\% & 14.07\% & 14.88\% \\
    \midrule
    
    \toprule
    \multirow{2}{*}{\textbf{Models}} & \multirow{2}{*}{\textbf{Metrics}} & \multicolumn{2}{c|}{\textbf{The last conv layer}} & \multicolumn{2}{c|}{\textbf{Last 14$\times$14 conv layer}} & \multicolumn{2}{c}{\textbf{Last 28$\times$28 conv layer}} \\
    \cmidrule{3-4} \cmidrule{5-6} \cmidrule{7-8} 
    & & \textbf{$\alpha=4$} & \textbf{$\alpha=8$} & \textbf{$\alpha=4$} & \textbf{$\alpha=8$} & \textbf{$\alpha=4$} & \textbf{$\alpha=8$} \\
    \midrule
    \multirow{3}{*}{\textit{ResNet50V2}} 
    & \textbf{AUC-D} $\uparrow$ 
    & \textbf{37.29\%} & 35.82\% & 36.36\% & 36.64\% & 32.28\% & 31.43\%  \\
    \cmidrule{2-8}
    &  Neg AUC $\uparrow$ 
    & \textbf{42.87\%} & 42.15\% & 41.95\% & 42.28\% & 38.25\% & 37.95\%  \\
    &  Pos AUC $\downarrow$ 
    & \textbf{5.58\%} & 6.33\% & 5.59\% & 5.64\% & 5.97\% & 6.52\%  \\
    \bottomrule
    \end{tabular}
\end{table*}

\subsection{Ablation Study for percentile $P$ selection}
We conduct a comprehensive ablation study to determine the optimal value for the percentile hyperparameter $P$, which governs the threshold applied to the analysis mask. Experiments are performed across two representative architectures—ViT-B16(transformer-based) and ResNet50V2(CNN-based)—with $P$ varied systematically from 0 to 95.

The quantitative results, summarized in Table \ref{tab8}, reveal a consistent trend across both architectures: performance, as measured by our primary evaluation metric AUC-D, improves monotonically with increasing $P$, reaching its optimum at $P=95$ for both models. This finding suggests that aggressive thresholding effectively suppresses background noise while preserving semantically meaningful activations.

\begin{figure}[!h]
\centering
\includegraphics[width=1\linewidth]{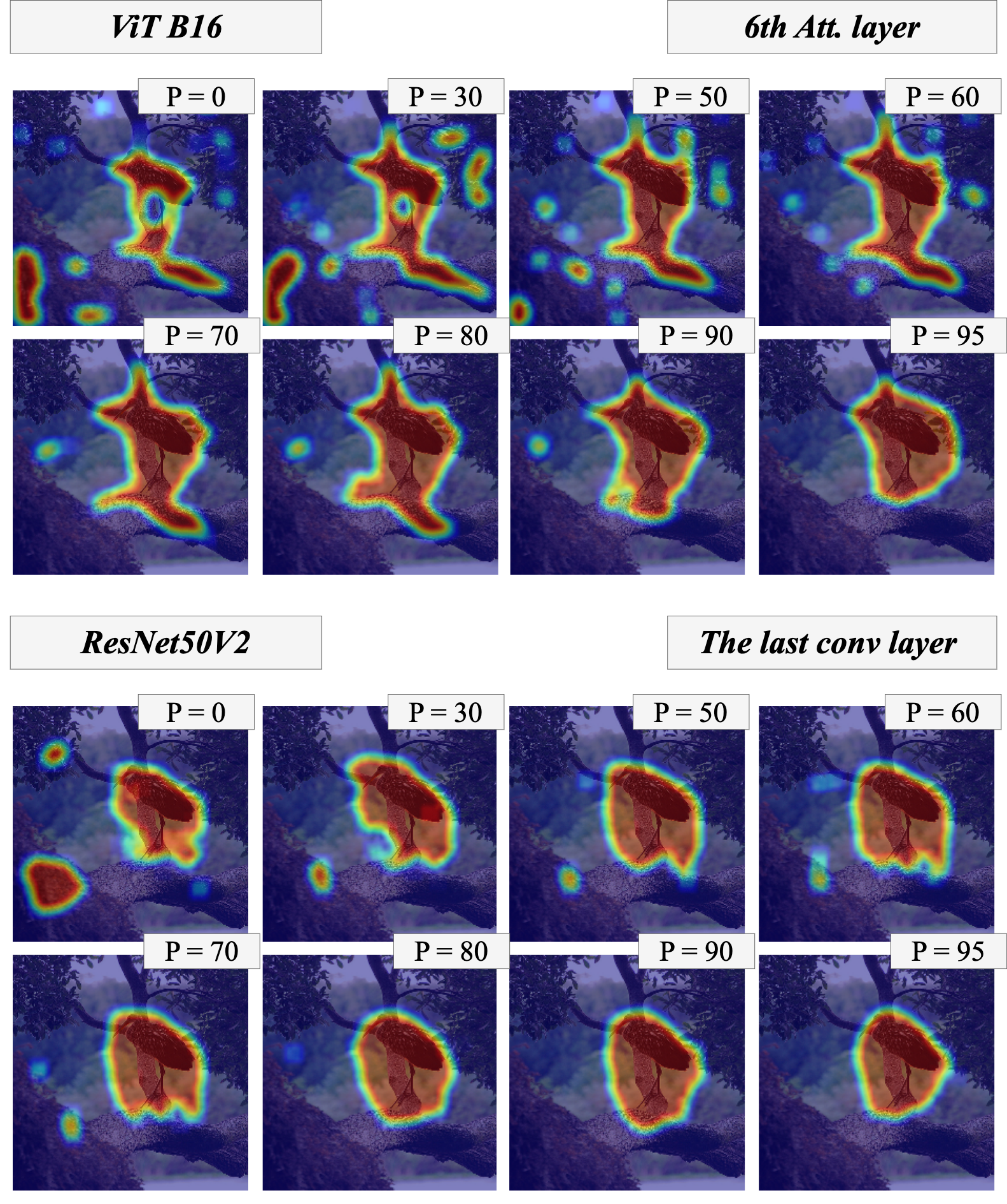}
    \caption{Qualitative ablation study on the percentile hyperparameter $P$. For both ViT-b16 and ResNet50V2, increasing $P$ leads to more accurate and less noisy object localization. This visually confirms that $P=95$ yields the most coherent saliency maps, aligning with the quantitative results in Table~\ref{tab8}.}
    \label{figA4}
\end{figure}

The qualitative analysis presented in Figure \ref{figA4} corroborates these quantitative observations. At lower values of $P$, the generated saliency maps exhibit scattered activations and considerable background noise. As $P$ increases, the maps become progressively more refined, with activations converging toward the object of interest. At $P=95$, both architectures produce the most coherent and semantically meaningful visual explanations, accurately delineating object boundaries with minimal spurious activations. Based on these consistent findings across both quantitative metrics and qualitative inspection, we adopt $P=95$ as the default setting for all subsequent experiments reported in this paper.

\begin{table*}[!t]
  \caption{Ablation study on a percentile $P$ for transformer(ViT) and CNN(ResNet). The percentile $P=95\%$ is the best. The best-performing configuration is highlighted in bold.}
  \label{tab8}
  \centering
  \begin{tabular}{l|cccccccc}
    \toprule
    \multicolumn{9}{c}{\textit{\textbf{ViT-b16}}} \\
    \midrule
    \multirow{2}{*}{\textbf{Metrics}} & \multicolumn{8}{c}{6th Att. layer and $\alpha=4$} \\
    \cmidrule{2-9}
    & \textbf{$P=0$} & \textbf{$P=30$} & \textbf{$P=50$} & \textbf{$P=60$} & \textbf{$P=70$} & \textbf{$P=80$} & \textbf{$P=90$} & \textbf{$P=95$} \\
    \midrule
    \textbf{AUC-D} $\uparrow$ & 12.00\% & 20.76\% & 24.49\% & 27.27\% & 30.23\% & 33.16\% & 35.76\% & \textbf{36.87\%} \\
    \midrule
    Neg AUC $\uparrow$ & 32.18\% & 37.71\% & 40.16\% & 42.06\% & 44.12\% & 46.28\% & 48.39\% & \textbf{49.29\%} \\
    Pos AUC $\downarrow$ & 20.18\% & 16.95\% & 15.67\% & 14.79\% & 13.90\% & 13.11\% & 12.62\% & \textbf{12.42\%} \\
    \midrule
    
    \toprule
    \multicolumn{9}{c}{\textit{\textbf{ResNet50V2}}} \\
    \midrule
    \multirow{2}{*}{\textbf{Metrics}} & \multicolumn{8}{c}{The last conv layer and $\alpha=4$} \\
    \cmidrule{2-9}
    & \textbf{$P=0$} & \textbf{$P=30$} & \textbf{$P=50$} & \textbf{$P=60$} & \textbf{$P=70$} & \textbf{$P=80$} & \textbf{$P=90$} & \textbf{$P=95$} \\
    \midrule
    \textbf{AUC-D} $\uparrow$ & 23.09\% & 24.67\% & 26.14\% & 27.02\% & 28.68\% & 30.22\% & 32.84\% & \textbf{37.29\%} \\
    \midrule
    Neg AUC $\uparrow$ & 34.19\% & 35.2\% & 35.8\% & 36.42\% & 37.5\% & 38.34\% & 39.99\% & \textbf{42.87\%} \\
    Pos AUC $\downarrow$ & 11.1\% & 10.53\% & 9.66\% & 9.4\% & 8.82\% & 8.12\% & 7.15\% & \textbf{5.58\%} \\
    \bottomrule
    \end{tabular}
\end{table*}

\subsection{Effectiveness of Individual Components}
To evaluate the intrinsic contribution of each core mechanism within the SCAN framework, we conducted a component-wise ablation study on the ImageNet using the ResNet50V2. As summarized in Table \ref{tab9}, the $\alpha=1$ setting, which does not distinguish between positive and negative regions, resulted in the most significant performance degradation, with the AUC-D score plummeting to 4.19\%. This underscores the critical role of $\alpha$ in enforcing selective feature reconstruction by assigning opportunity costs to information-rich regions. Furthermore, removing the Gradient Mask led to a decrease in AUC-D to 23.09\%, confirming its necessity for filtering class-discriminative representations from intermediate layers. Omitting the Gaussian Blur target $\tilde{Y}$ also caused the performance to drop to 12.76\%, as the model struggles to reconstruct high-frequency details lost during downsampling. Lastly, replacing the proposed stretching sine loss with a sigmoid loss resulted in a score of 32.63\%. These results validate that the synergy of the proposed components is essential for achieving optimal visual explanation performance.


\begin{table}[!h]
  \caption{Ablation study of SCAN components on ImageNet using ResNet50v2. The results demonstrate the impact on performance when key modules are removed or replaced, validating the contribution of each component.}
  \label{tab9}
  \centering
  \begin{tabular}{l|r|rr}
    \toprule
        \textbf{Components} & \textbf{AUC-D} $\uparrow$ & \textbf{Neg AUC} $\uparrow$ & \textbf{Pos AUC} $\downarrow$  \\
    \midrule
        $\alpha=1$
        & 4.19\% & 25.82\% & 21.63\% \\
        w/o Gradient Mask
        & 23.09\% & 34.19\% & 11.1\% \\
        w/o Gaussian Blur
        & 12.76\% & 29.42\% & 16.66\% \\
        $\sin \rightarrow \sigma$ for Conf. loss
        & 32.63\% & 40.13\% & 7.5\% \\
    \midrule
        \textbf{SCAN} 
        & \textbf{37.29\%} & \textbf{42.87\%} & \textbf{5.58\%}  \\
    \bottomrule
  \end{tabular}
\end{table}


\section{Discussions and Conclusions}

In this study, we devised SCAN, a novel visual explanation framework designed to address the critical trade-offs between architecture-specific and universal explanation methods. By reconstructing internal feature representations and generating a self-confidence map guided by the information bottleneck theory, SCAN provides high-fidelity, feature-rich explanations applicable to both CNN and transformer.

Quantitatively, SCAN demonstrated robust generalizability across diverse datasets. Notably, on the CUB dataset, it achieved a superior performance, topping key metrics such as AUC‑D and Negative AUC, while maintaining low Positive AUC. Qualitatively, SCAN generates significantly clearer and more precise explanations compared to existing methods. While other approaches often produce diffuse or fragmented heatmaps, SCAN accurately delineates object boundaries. Together, these quantitative and qualitative results confirm the effectiveness of the proposed method.

Despite these promising results, we acknowledge several limitations that remain open avenues for future research. The current framework requires separate training for the analysis network, which introduces computational overhead. However, in practice, visual explanations at inference time are extremely rapid, and only one trained analysis network is sufficient for a target network.

In conclusion, SCAN represents a significant step toward a unified and more interpretable XAI framework. By bridging the gap between the high fidelity of model-specific techniques and the broad applicability of universal ones, SCAN enhances the transparency and reliability of deep learning models, paving the way for more trustworthy and understandable AI systems. To support ongoing research, we have made the SCAN code available at: 

\url{https://github.com/gompanghee/SCAN}



\bibliographystyle{IEEEtran}
\bibliography{reference}

\begin{IEEEbiography}[{\includegraphics[width=1in,height=1.25in,clip,keepaspectratio]{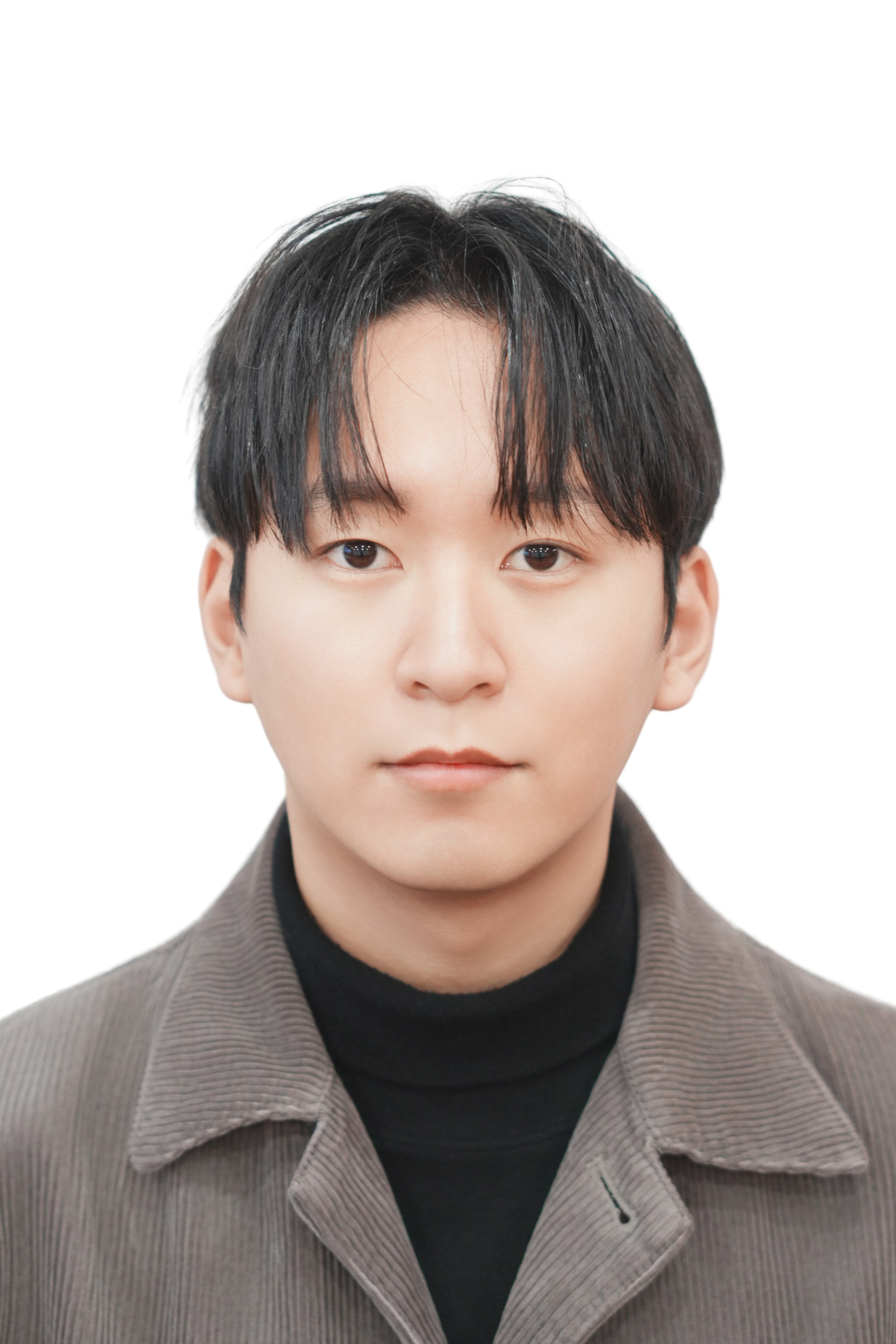}}]{Gwanghee Lee}{\space}has been in progress of Ph.D. course in Computer Science at Chungnam National University since 2021. His research interests are Vision-Language Model, Visual Explanation, Explainable AI, Generalization, Robustness, Long-tailed Recognition, Fine-Grained Classification, Face Alignment, Pose Estimation, Keypoint Detection, Computer Vision, and Deep Learning.
\end{IEEEbiography}

\begin{IEEEbiography}[{\includegraphics[width=1in,height=1.25in,clip,keepaspectratio]{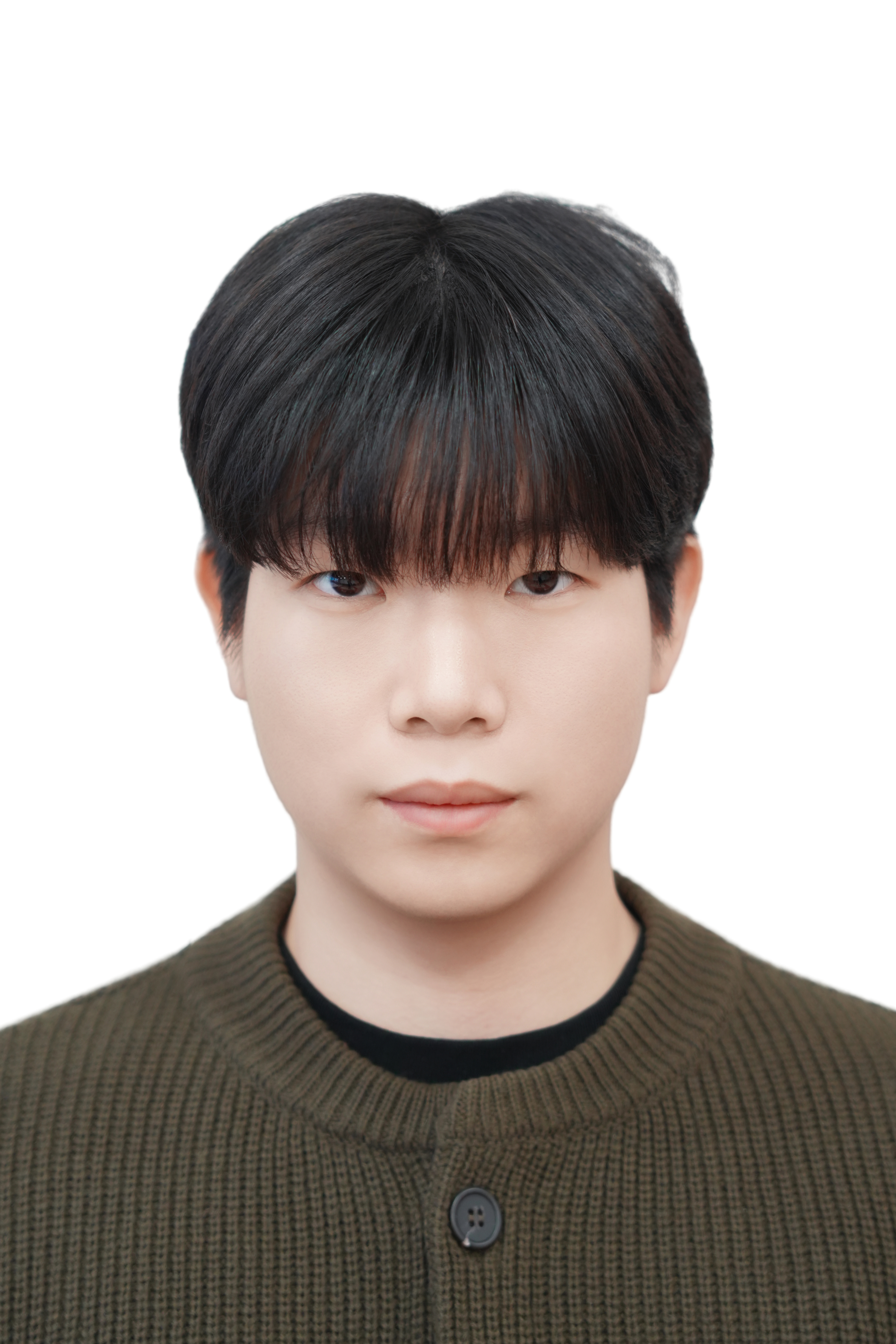}}]{Sungyoon Jeong}{\space}is currently pursuing the B.S. degree in artificial intelligence with the Department of Artificial Intelligence, Chungnam National University, Daejeon, South Korea. His research interests include Computer Vision and Deep Learning, with a specific focus on Explainable AI (XAI), Generalization, Robustness and Long-tailed Recognition of neural networks.
\end{IEEEbiography}

\begin{IEEEbiography}[{\includegraphics[width=1in,height=1.25in,clip,keepaspectratio]{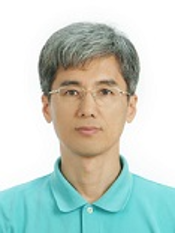}}]{Kyoungson Jhang}{\space}received B.S., M.S., and Ph.D. degrees in Department of Computer Engineering from Seoul National University in 1986, 1988, and 1995, respectively. Since September 2001, he has been working as a professor for the Department of Computer Science and Engineering at Chungnam National University, Daejeon, Korea. His research focuses on Computer Vision and Deep Learning
\end{IEEEbiography}

\end{document}